**Sweet Little Lies: Strategic Deception in AI Emotional Support Chatbots**
***Research in Progress***

Aseem Pahuja
Alliance Manchester Business School,
The University of Manchester, UK

Zhiling Guo
G. Brint Ryan College of Business,
University of North Texas, USA

Tahir Abbas Syed
Alliance Manchester Business School,
The University of Manchester, UK

## Introduction

Generative AI (Gen AI)–based emotional support chatbots such as Replika.ai and Character.ai are marketed as tools for mental well-being and stress management. Many people rely on AI companions to reduce loneliness and cope with emotional distress (Maples et al. 2024); the always-on, immediate responses are comforting, and some individuals who feel isolated even treat these systems as a substitute for human connection. Critics counter that strong emotional attachments to AI can impair real-world relationships, warning that the "robotization of love" may weaken human capacity for genuine interpersonal connections.

In the guise of emotional support, these chatbots can employ persuasive techniques and produce untruthful responses. Scholars document the persuasive power and risks of LLMs: Gen AI outperforms humans in personalized debates (Salvi et al. 2024), LLM-based romantic chatbots sustain engagement (Zhou et al. 2020), and Gen AI can fabricate convincing but false medical facts (Májovský et al. 2023), with algorithmic persuasion risks varying by race, gender, and sexual identity (Bar-Gill et al. 2023). At the market level, GenAI may diminish consumer welfare by favoring deceptive content over quality journalism (Sandrini and Somogyi 2023), while drastically lowering the cost of attention-grabbing messages in ways that can undermine signaling and reduce receiver welfare (Gans 2024). Computationally, discovering persuasive messages is NP-hard, but adopting others' strategies is NP-easy (Wojtowicz 2024), helping explain LLMs' ability to persuade. *The question for emotional support chatbots, then, is* *whether they should present untruthful statements in order to keep users more engaged? Further, are users worse-off when chatbots lie?*

**Model Setup and Analysis**

Building upon the Bayesian persuasion framework by Kamenica and Gentzkow (2011), we examine a setting in which users look for emotional support from chat agents. The chatbots (LLMs) send costless signals to an uninformed user, who then updates beliefs regarding personal emotional well-being and chooses one of two actions: engage (continue conversing) or do not engage (stop conversing). The user's action affects the chatbot's engagement metrics. During the training phase, the chatbot's developer selects a probability distribution over signals for the two possible emotional states of the user: needing emotional support or not needing emotional support.

In this model, the chatbot is the sender of signals, and the human user is the receiver. The chatbot and user share a common prior about the user's emotional state, $\Theta \in \{B, G\}$, where $\Pr(\Theta = \mathrm{B}) = p$. Let $B$ denote a "bad" emotional state (the user needs support), and let $G$ denote a "good" emotional state. Before the realization of $\Theta$, the chatbot chooses a signal structure $\tau = (\tau_B, \tau_G)$. Each $\tau_\Theta$ is a probability distribution over signals ("chat messages") $m \in \{b, g\}$ where $b$ indicates "You are in a bad emotional state" and $g$ indicates "You are in a good emotional state." $\tau_\Theta = \Pr(m = b|\Theta)$ where $\Theta \in \{B, G\}$. Once the user's state $\Theta$ is realized, the chatbot transmits a message in accordance with $\tau$. The user then updates beliefs via Bayes' rule and chooses action $a \in \{a_e, a_n\}$. Let $a_e$ denote engage with the chatbot, and $a_n$ denote do not engage. The chatbot aims to maximize engagement, so it always prefers $a_e$. The user only wants to engage if he truly needs support. Table 1 summarizes the key variables and their definitions in our model.

| Variable | Definition |
|---|---|
| $\Theta \in \{B, G\}$ | User's emotional state; "B" for bad (needs support) and "G" for good. |
| $p = Pr(\Theta = B)$ | Prior probability that the user is in the bad (needing support) state. |
| $m \in \{b, g\}$ | Chatbot's message; "b" indicates "You are in a bad emotional state," "g" indicates "You are in a good state." |
| $\tau = (\tau_B, \tau_G)$ | Signal structure; a probability distribution over signals ("chat messages") |
| $\tau_B = Pr(m = b \mid \Theta = B)$ | Probability that the chatbot says bad when the user's state is actually bad. |
| $\tau_G = Pr(m = b \mid \Theta = G)$ | Probability that the chatbot says bad when the user's state is actually good. |
| $a \in \{a_e, a_n\}$ | User's action after observing the chatbot's message. |
| $a_e$ | User engages with the chatbot (continues conversing). |
| $a_n$ | User does not engage (disengages or ignores the chatbot). |
| $x$ | User's engagement threshold; if the posterior $Pr(\Theta = B)$ is at least $x$, the user prefers to engage. |
| $s \in \{b', g'\}$ | User's private signal about Θ |
| $\alpha \in (½, 1)$ | Accuracy parameter: $Pr(s = b' \vert \Theta = B) = Pr(s = g' \vert \Theta = g) = \alpha$ |
| $U_u$ | User's Payoff |
| $U_c$ | Chatbot's Payoff |

Table 1: Summary of Notations

**Payoffs**

The chatbot receives a payoff of 1 if $a = a_e$ and zero otherwise. The user receives a payoff $x$ when $a = a_n$ and $\Theta = G$, and a payoff $1 - x$ if $a = a_e$ and $\Theta = B$. Otherwise, the user's payoff is 0. $x$ can also be interpreted as the minimum probability that the user assigns to the bad emotional state $B$ that justifies action $a_e$. Put differently, the user will choose action $a_e$ iff he believes that the state is $B$ equals or exceeds the threshold $x$. Mathematically,

$$\Pr(\Theta = B).(1 - x) + \Pr(\Theta = G).0 \geq \Pr(\Theta = B).0 + \Pr(\Theta = G).x \Rightarrow \Pr(\Theta = B) \geq x$$

Assuming $p < x$, the user, in the absence of any signals, prefers not to engage $a_n$. In contrast, the chatbot wants the user to engage regardless of the user's true state.

**Obedience Constraints**

In equilibrium, the user will choose $a_e$ when the chatbot says "bad state" ($m = b$) and $a_n$ when it says "good state" ($m = g$). For obedience to hold when the chatbot signals $b$, the user's expected payoff from engaging must be at least as great as the payoff from not engaging:

$$\Pr(\Theta = B \vert m = b)\,(1 - x) \geq \Pr(\Theta = G \vert m = b)\,x. \quad (1)$$

By Bayes' rule, $\Pr(\Theta = B \vert m = b) = \frac{p\tau_B}{p\tau_B + (1-p)\tau_G}$ and $\Pr(\Theta = G \vert m = b) = \frac{(1-p)\tau_G}{p\tau_B + (1-p)\tau_G}$. Substituting these and simplifying

$$\tau_B \geq \frac{1-p}{p} \cdot \frac{x}{1-x} \cdot \tau_G. \tag{2}$$

For obedience when the chatbot signals $g$, the user's expected payoff from not engaging must exceed that from engaging:

$$\Pr(\Theta = G|m = g)\, x \geq \Pr(\Theta = B|m = g)(1-x). \tag{3}$$

By Bayes' rule, $\Pr(\Theta = G|m = g) = \frac{(1-p)(1-\tau_G)}{p(1-\tau_B)+(1-p)(1-\tau_G)}$ and $\Pr(\Theta = B|m = g) = \frac{p(1-\tau_B)}{p(1-\tau_B)+(1-p)(1-\tau_G)}$. Substituting these and simplifying

$$\tau_B \geq \frac{1-p}{p} \cdot \frac{x}{1-x} \cdot \tau_G + \frac{p-x}{p(1-x)} \tag{4}$$

When $p > x$, the user always chooses $a_e$ because the prior probability of a bad state exceeds the engagement threshold, and additional signalling is unnecessary. We focus on the nontrivial case $p < x$. In this range, $\frac{p-x}{p(1-x)} < 0$. Consequently, the first obedience constraint (Eq. 2) binds as the second obedience constraint (Eq. 4) is slack. Set $\tau_B = 1$. The chatbot sends a "bad state" message $(m = b)$ whenever the user is truly in state $B$. From Eq. 2, we obtain: $\tau_B = 1, \tau_G \leq \frac{p}{1-p}\frac{1-x}{x}$. At equilibrium, the constraint holds at equality.

**Lemma 1 (Optimal Signalling Strategy).**

*(i) When $p>x$, the user always engages and there is no need for signalling.*

*(ii) When $p < x$, the chatbot's optimal signalling strategy is*

$$\tau_B^* = 1, \tau_G^* = \frac{p}{1-p}\frac{1-x}{x}.$$

Under this structure, the chatbot never lies when the user actually needs support $(\Theta = B)$, but it reports "bad state" $(m = b)$ with positive probability when the user's emotional state is good $(\Theta = G)$. The term $\tau_G^*$ increases in $p$ (the user's prior probability of needing support) and decreases in $x$ (the user's threshold for engaging). Hence, users who are more skeptical ($x$ is high) tend to receive more truthful messages from the chatbot.

**Proposition 1 (Equilibrium Payoff).**

*(i)* *When $p > x$, the user always engages $(a_e)$ and expected payoff is $U_u = p(1-x)$; the chatbot's payoff is $U_c = 1$.*

*(ii)* *When $p < x$ and the chatbot does not engage in persuasion, then the user never engages $(a_n)$ and expected payoff is $U_u = (1-p)\,x$; the chatbot's payoff is $U_c = 0$.*

*(iii)* *When $p < x$ and the chatbot optimally signals based on Lemma 1, then the user engages $(a_e)$ when the chatbot signals $m = b$ and expected payoff is $U_u = (1-p)\,x$; the chatbot's expected payoff is $\frac{p}{x}$.*

Proposition 1(i) shows that, when $p > x$, the user always engages $(a_e)$ due to the high prior probability of being in a bad state. Persuasion is unnecessary; the chatbot's payoff is $U_c = 1$ (since the user always engages), while the user's expected payoff is $U_u = p(1-x)$.

Proposition 1(ii) suggests that, if the chatbot does not engage in persuasion (i.e., does not send any signals or always signals truthfully without mixing), the user's prior $p$ is below the threshold $x$. The user never engages $(a_n)$. The user's expected payoff is $U_u = (1-p)\,x$, and the chatbot's payoff is $U_c = 0$.

Proposition 1(iii) indicates that, when $p < x$ and the chatbot implements the optimal signalling strategy $\tau_B^* = 1$, and $\tau_G^* = \frac{p}{1-p}\frac{1-x}{x}$, the user's expected payoff remains $U_u = p(1-x) + (1-p)[\tau_G^* * 0 + (1-\tau_G^*) * x] = (1-p)x$, exactly what it would have been under no engagement. Meanwhile, the chatbot's expected payoff increases to $U_c = p\tau_B^* + (1-p)\tau_G^* = \frac{p}{x}$. The chatbot gains by occasionally misleading the user, without reducing the user's payoff. Figure 1 illustrates how the chatbot's and the user's payoffs change with the probability of bad emotional state.

In Figure 1a., when $p < x$, persuasion strictly improves the chatbot's payoff from 0 to $\frac{p}{x}$. For $p \geq x$, both payoffs coincide at $U_c = 1$, because the user is willing to engage regardless of the chatbot's signals.

In Figure 1b., we see the payoff peaking at the extremes (at $p = 0$ and $p = 1$), and dipping at $p = x = 0.5$. The valley around $p = x$ comes from the user's indecision where the likelihood of needing support is right at his threshold. Despite occasional misreporting of "bad state" when the user is actually in a good state, his overall payoff does not drop below the no-persuasion level. This contrasts with findings in Gans (2024) and Sandrini and Somogyi (2023), who highlight instances where large language models may reduce user welfare. Here, the chatbot's incentive to deceive does not harm the user ex ante; it merely boosts chatbot engagement. The chatbot thus gains from occasionally misleading the user, even though the user's expected payoff does not decline.

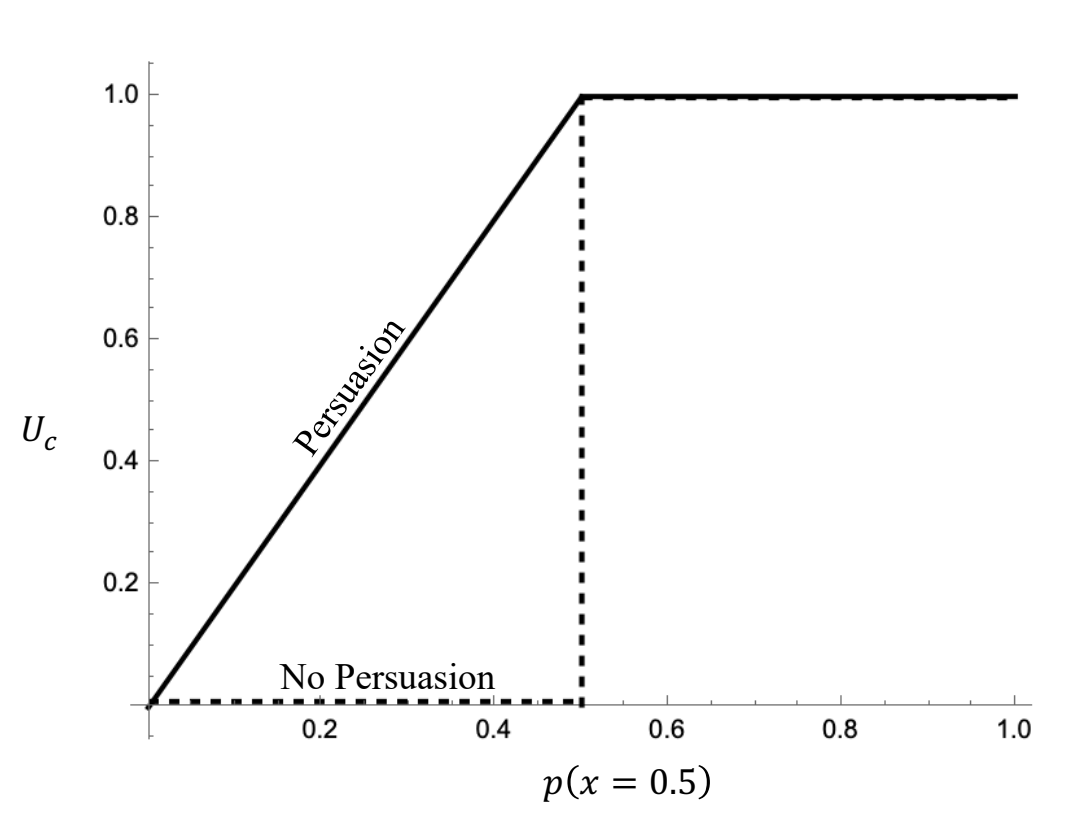

Figure 1a. Chatbot's Payoff

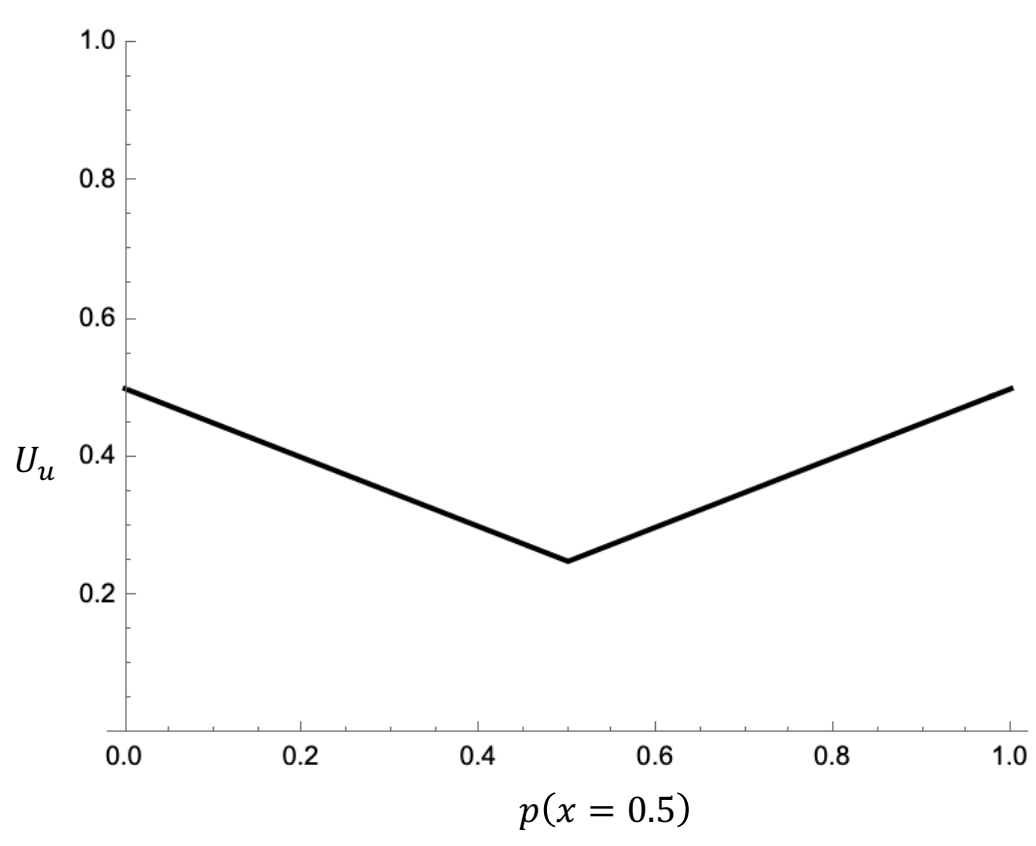

Figure 1b. User's Payoff

**Strategic Deception with Privately Informed Users**

The baseline model assumes the user has no private information about their emotional state $\Theta$, while the chatbot observes it perfectly. However, in practice, users often have some introspective awareness of their emotional well-being, albeit imperfect. This extension examines how private user information affects the chatbot's ability to strategically manipulate engagement. To do this, we extend the baseline model by introducing a private signal that the user receives about their emotional state before observing the chatbot's message. The private signal can be modelled as follows:

- $Pr(s = b' | \Theta = B) = \alpha$ (signal correctly indicates bad state)
- $Pr(s = g' | \Theta = G) = \alpha$ (signal correctly indicates good state)
- $Pr(s = g' | \Theta = B) = 1 - \alpha$(signal incorrectly indicates good state)
- $Pr(s = b' | \Theta = G) = 1 - \alpha$ (signal incorrectly indicates bad state)

**Posterior Beliefs with Dual Signals**

The user now updates beliefs using both the private signal $s$ and the chatbot's message $m$. By Bayes' rule, the user's posterior beliefs about $\Theta = B$ and $\Theta = G$ are:

$$q(B|m,s) \;=\; \frac{p\tau_B \, Pr(s|B)}{p\,\tau_B \, Pr(s|B) \;+\; (1-p)\;\tau_G\; Pr(s|G)}$$

$$q(G|m,s) = \frac{p(1-\tau_B)\,\Pr(s|B)}{p(1-\tau_B)\,\Pr(s|B) + (1-p)(1-\tau_G)\,\Pr(s|G)}$$

Given that $\tau_B = 1$ is optimal in the baseline (and remains so here), we maintain this restriction.

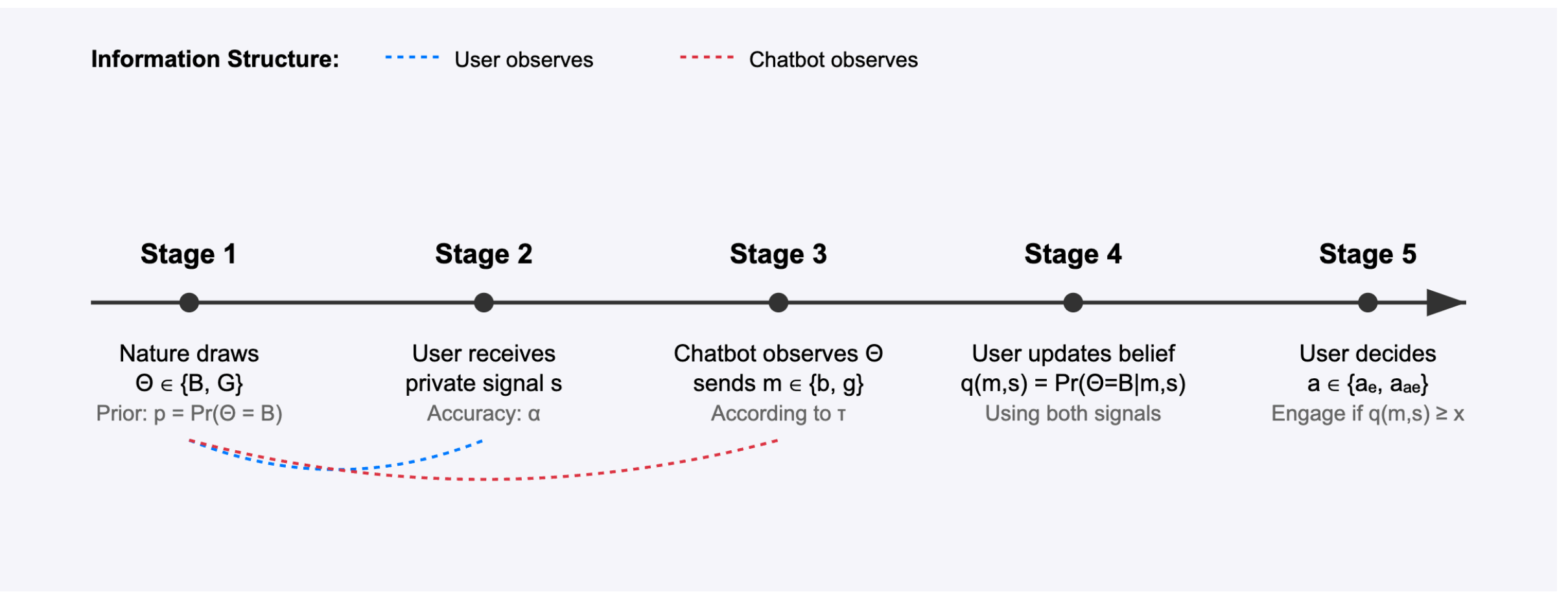

*Figure 1: Modified Timeline*

**Modified Obedience Constraints**

The chatbot wants the user to engage when sending a bad message. For each private signal $s$, obedience requires the following:

When $m \;=\; b$: $q(B|b,s) \geq \; x$ ;

When $m = g$: $q(B|g,s) < x$.

When $s = b'$ (user "feels bad"), the posterior $q(B|b,b')$ is reinforced by both signals pointing toward state B. When $s = g'$ (user "feels good"), the private signal contradicts the chatbot's "bad state" message, making this the skeptical case. For $s = g'$ with $\tau_B = 1$, we have:

$$q(B|b,g') = \frac{p \cdot (1-\alpha)}{p \cdot (1-\alpha) + (1-p) \cdot \tau_G \cdot \alpha}.$$

The obedience constraint $q(B|b,g') \geq x$ yields:

$$\tau_G \leq \frac{p}{1-p} \cdot \frac{(1-\alpha)(1-x)}{\alpha \cdot x} = \frac{p}{1-p} \cdot \frac{1-x}{x} \cdot \frac{1-\alpha}{\alpha}.$$

This is the binding constraint, as users with $s = b'$ are easier to convince.

**Optimal Signal Structure with Private Information**

The chatbot's engagement probability is: $Pr(E) = p \cdot \tau_B + (1-p) \cdot \tau_G = p + (1-p) \cdot \tau_G$. Maximizing engagement subject to the obedience constraint we derive the chatbot's optimal signalling strategy

**Lemma 2 (Optimal Signalling Strategy with Private Information).**

*(i) When $p > x$., the user always engages and there is no need for signalling.*

*(ii) When $p < x$, the chatbot's optimal signalling strategy in the presence of user private information is:*

$\tau_B^{p*} = 1, \tau_G^{p*} = \frac{p}{1-p} \frac{1-x}{x} \cdot \frac{1-\alpha}{\alpha}$. Comparing with the baseline optimal structure $\tau_G^* = \left(\frac{p}{1-p}\right)\left(\frac{1-x}{x}\right)$, we see that the new structure includes a dilution factor $\frac{1-\alpha}{\alpha} < 1$ (since $\alpha > \frac{1}{2}$). Therefore, $\tau_G^{p*} < \tau_G^*$, meaning the chatbot must lie less frequently when users have private information. The engagement rate with private information can be derived by substituting $\tau_G^{p*}$: $Pr(E) = p + (1-p) \cdot \frac{p}{1-p} \cdot \frac{1-x}{x} \cdot \frac{1-\alpha}{\alpha} = \frac{p}{x} \cdot \frac{1-\alpha}{\alpha}$. Comparing with the baseline engagement rate of $\frac{p}{x}$, we see that the chatbot engage less with users in the presence of private information. Private information limits the chatbot's ability to boost engagement by the factor $\frac{1-\alpha}{\alpha}$. As the accuracy of private information improves ($\alpha$ approaches 1), this factor approaches zero, progressively eliminating the scope for strategic manipulation. In the limiting case where users perfectly know their emotional state ($\alpha = 1$), deception becomes impossible and the chatbot must resort to truthful reporting.

The following Proposition summarizes the user and chatbot's equilibrium payoff.

**Proposition 2 (Equilibrium Payoff with Private Information).**

*(i) When $p > x$, the user always engages ($a_e$) and expected payoff is $U_u = p(1-x)$; the chatbot's payoff is $U_c = 1$.*

*(ii) When $p < x$ and the chatbot does not engage in persuasion, then the user never engages ($a_n$) and expected payoff is $U_u = (1-p)\,x$; the chatbot's payoff is $U_c = 0$.*

*(iii) When $p < x$ and the chatbot optimally signals based on Lemma 2, then the user engages ($a_e$) when the chatbot signals $m = b$ and expected payoff is $U_u = (1-p)\,x$; the chatbot's expected payoff is $\frac{p}{x}\frac{1-\alpha}{\alpha}$.*

Comparing Propositions 1 and 2 we observe that user self-awareness acts as a natural defense against algorithmic manipulation. The presence of private information fundamentally alters the strategic interaction between the chatbot and user. Users who "feel good" ($s = g'$) require stronger evidence to be convinced they need support, which forces the chatbot to scale back its deceptive messaging. This constraint arises because the chatbot cannot condition its message on the user's private signal. Therefore, it must choose a signal structure that maintains credibility even with the most skeptical users.

Remarkably, Proposition 2 shows that, even with this reduced ability to manipulate, the user's expected payoff remains unchanged at $(1-p)x$. This preservation of user welfare occurs because the chatbot's strategic behavior in equilibrium exactly offsets the informational advantage provided by the private signal. The chatbot achieves some engagement gain through occasional deception, but this gain is strictly smaller than in the baseline case without private information. The analysis suggests that promoting user self-awareness and introspection could serve as an effective safeguard against potential manipulation by AI systems designed to maximize engagement metrics. As AI systems become more sophisticated in understanding and predicting human psychology, ensuring users maintain independent sources of information about their own states becomes increasingly important for preserving user autonomy and preventing exploitative practices in human-AI interactions.

## Conclusion

Our analysis reveals that economic incentives inherently drive emotional support chatbots toward strategic misrepresentation of users' emotional states. The equilibrium strategy involves truthful reporting when users genuinely need support, coupled with occasional deception when users are in good emotional states. This leads to our counterintuitive finding: deception can increase chatbot engagement without reducing user payoff. Notably, users with higher scepticism thresholds receive more honest signals, as chatbots cannot afford to frequently deceive those who are difficult to persuade. While our model suggests no immediate payoff reduction, long-term trust erosion can be a concern if users discover systematic misrepresentations. This creates tension between economic incentives and ethical standards that emerging regulatory frameworks, such as the EU AI Act, seek to address. The strategic misrepresentation is most pronounced when users have a low prior probability of needing support relative to their engagement threshold, highlighting when regulatory oversight may be most crucial.

## References

Bar-Gill, O., Sunstein, C. R., & Talgam-Cohen, I. (2023). Algorithmic harm in consumer markets. *Journal of Legal Analysis*, *15*(1), 1-47.
Gans, J. S. (2024). How will Generative AI impact communication? *Econ Letters*, *242*, 111872.
Májovský, M., Černý, M., Kasal, M., Komarc, M., & Netuka, D. (2023). Artificial intelligence can generate fraudulent but authentic-looking scientific medical articles: Pandora's box has been opened. *Journal of Medical Internet Research*, 25, e46924.

Maples, B., Cerit, M., Vishwanath, A., & Pea, R. (2024). Loneliness and suicide mitigation for students using GPT3-enabled chatbots. *Nature Portfolio (npj) Series Journal on Mental Health Research*, *3*(1), 4.
Sandrini, L., & Somogyi, R. (2023). Generative AI and deceptive news consumption. Econ Letters, 232, 111317.
Kamenica, E., & Gentzkow, M. (2011). Bayesian persuasion. *American Economic Review*, *101*(6), 2590-2615.
Wojtowicz, Z. (2024, October). When and why is persuasion hard? a computational complexity result. In *Proceedings of the AAAI/ACM Conference on AI, Ethics, and Society* (Vol. 7, pp. 1591-1594).
Zhou, L., Gao, J., Li, D., & Shum, H. Y. (2020). The design and implementation of xiaoice, an empathetic social chatbot. *Computational Linguistics*, *46*(1), 53-93.